\documentclass[10pt,twocolumn,letterpaper]{article}

\usepackage{wacv}
\usepackage{amsmath}
\usepackage{algorithmic}
\usepackage{array}
\usepackage[caption=false,font=normalsize,labelfont=sf,textfont=sf]{subfig}
\usepackage{textcomp}
\usepackage{stfloats}
\usepackage{url}
\usepackage[font={small,bf}]{caption}
\usepackage{graphicx}
\usepackage{bm}
\usepackage{diagbox}
\usepackage{ctable}
\usepackage{booktabs}
\usepackage{mathrsfs}
\usepackage{tocloft} 
\usepackage{changepage}
\usepackage{multirow}
\usepackage{adjustbox}
\usepackage{makecell}
\usepackage{xcolor} 
\usepackage{colortbl}   
\usepackage{tabularray}
\usepackage[accsupp]{axessibility}

\definecolor{best}{rgb}{0.86, 0.5, 0.5}
\definecolor{second}{rgb}{0.98, 0.78, 0.57}
\definecolor{third}{rgb}{1.0, 1.0, 0.56}

\usepackage[capitalize]{cleveref}
\crefname{section}{Sec.}{Secs.}
\Crefname{section}{Section}{Sections}
\Crefname{table}{Table}{Tables}
\crefname{table}{Tab.}{Tabs.}

\begin{document}

\title{NeFF-BioNet: Crop Biomass Prediction from Point Cloud \\ to Drone Imagery}
\author{
Xuesong Li$^{1,3}$, 
Zeeshan Hayder$^{2}$, 
Ali Zia$^{1,3}$, 
Connor Cassidy$^{1}$, 
Shiming Liu$^{1}$, 
Warwick Stiller$^{1}$, \\
Eric Stone$^{3}$, 
Warren Conaty$^{1}$, 
Lars Petersson$^{2}$, 
Vivien Rolland$^{1}$
 \\
$^{1}$CSIRO Agriculture and Food, $^{2}$CSIRO Data61, $^{3}$Australian National University, Australia \\ 
{\tt\small xuesong.li@csiro.au}
}

\twocolumn[{
\maketitle
\begin{center}
    \captionsetup{type=figure}
    \includegraphics[width=0.9\textwidth]{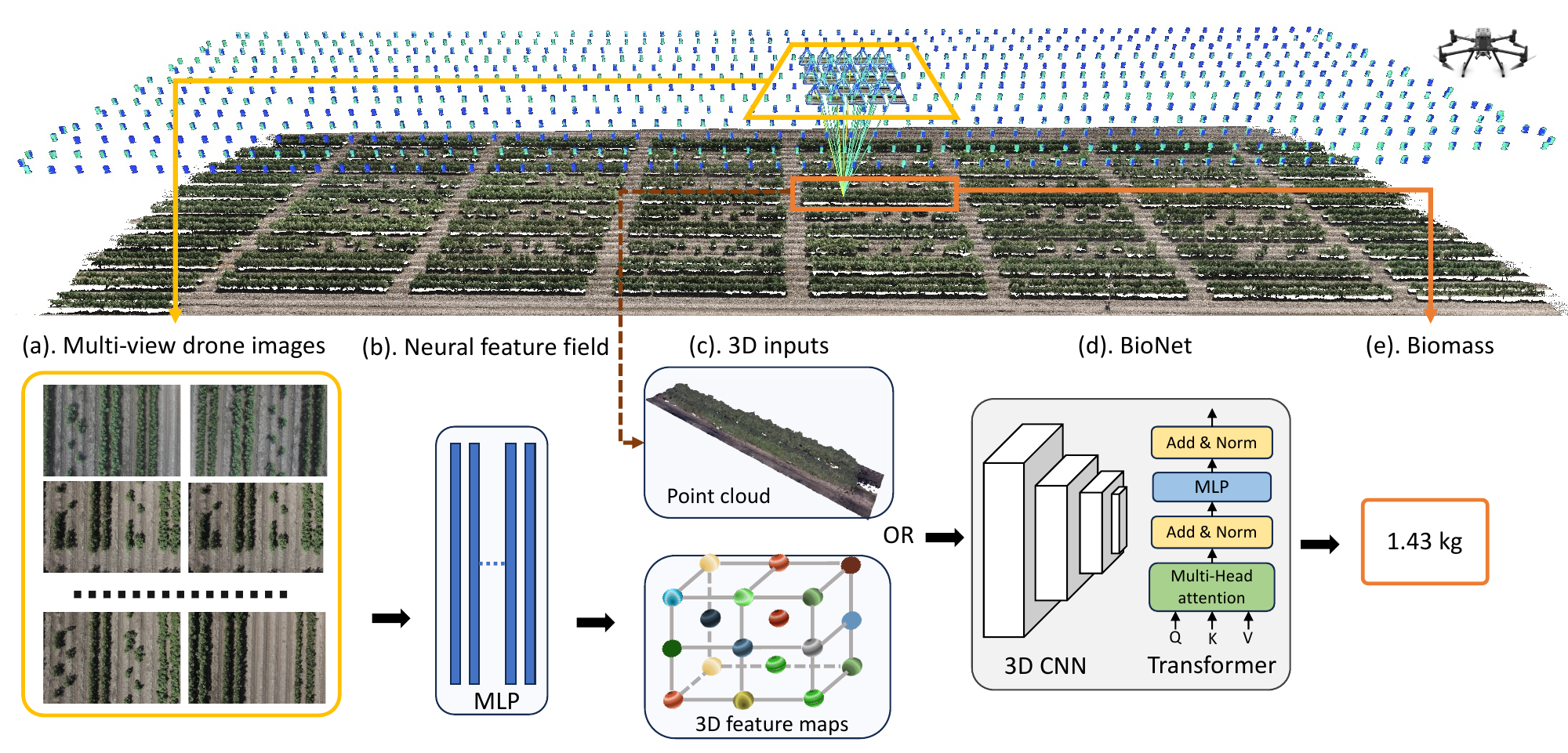}
    \setlength{\belowcaptionskip}{-4pt}
    \vspace*{-0.2cm}
    \captionof{figure}{Overview of our biomass prediction network: Our BioNet (d), consisting of a sparse 3D CNN backbone network and a transformer encoder, takes point cloud or 3D semantic feature maps as input and predicts biomass (e). For point cloud modality, the point cloud can be directly piped into BioNet as 3D inputs (c). For RGB imagery modality, a collection of drone images (a) above crops are gathered and processed by our NeFF (b) to generate 3D feature maps from 2D images as 3D inputs (c).}
    \label{fig:overal_framework}
\end{center}
}]

\begin{abstract}
Crop biomass offers crucial insights into plant health and yield, making it essential for crop science, farming systems, and agricultural research. However, current measurement methods, which are labor-intensive, destructive, and imprecise, hinder large-scale quantification of this trait. To address this limitation, we present a biomass prediction network (BioNet), designed for adaptation across different data modalities, including point clouds and drone imagery. Our BioNet, utilizing a sparse 3D convolutional neural network (CNN) and a transformer-based prediction module, processes point clouds and other 3D data representations to predict biomass. To further extend BioNet for drone imagery, we integrate a neural feature field (NeFF) module, enabling 3D structure reconstruction and the transformation of 2D semantic features from vision foundation models into the corresponding 3D surfaces. For the point cloud modality, BioNet demonstrates superior performance on two public datasets, with an approximate 6.1\% relative improvement (RI) over the state-of-the-art. In the RGB image modality, the combination of BioNet and NeFF achieves a 7.9\% RI. Additionally, the NeFF-based approach utilizes inexpensive, portable drone-mounted cameras, providing a scalable solution for large field applications.
\end{abstract}

\vspace*{-\baselineskip}

\section{Introduction}
\label{sec:intro}

The amount of organic matter a plant produces, known as biomass, is a key indicator of plant growth and productivity. Estimating biomass on a large scale (up to global maps) is a well-researched area within the remote sensing community~\cite{bullock2023estimating}, typically focusing on vegetation biomass at resolutions between 30m to 1km. This study targets high-resolution crop biomass estimation (sub-meter scale), with a focus on above-ground biomass, excluding root mass. Biomass serves as a vital trait in agricultural systems, aiding in monitoring plant establishment, informing decisions on replanting, and guiding various management actions such as chemical sprays. Additionally, it plays a crucial role in crop breeding programs for selecting test lines. However, large-scale measurement of biomass remains impractical due to the limitations of current methods~\cite{jimenez2018high}.

Traditional biomass quantification involves harvesting and measuring biomass over small areas (e.g., a square meter) and scaling these values to represent larger plots. This approach is destructive, labor-intensive, and lacks scalability, limiting its ability to generate temporal data. In contrast, computer vision techniques enable non-destructive, time-series biomass measurement~\cite{caballer2021prediction,sun2018field, ten2019biomass, pan2022biomass}. Specifically, LiDAR has been used to capture plant height and point density as proxies for biomass estimation, but its 3D feature limitations and self-occlusion reduce accuracy. To address this, deep learning methods have emerged, enabling more accurate 3D feature extraction from point clouds~\cite{pan2022biomass, li2024mmcbe}. However, these methods still suffer from limited training data and feature extraction challenges. To tackle these issues, we propose a biomass backbone network, built primarily on sparse 3D CNN operations over point clouds. This network benefits from translation invariance and parameter sharing, enhancing adaptation across smaller datasets. To capture global 3D features, we incorporate transformer-based attention mechanisms~\cite{vaswani2017attention}, which extract global representations from the 3D CNN feature maps for improved biomass predictions, as illustrated in~\cref{fig:overal_framework}. We evaluate BioNet on two public datasets~\cite{pan2022biomass, li2024mmcbe}, achieving superior prediction accuracy compared to the state-of-the-art (SOTA).

While high-resolution LiDAR data could limit scalability, vision-based biomass measurement methods present a promising alternative due to the widespread availability of low-cost cameras. To extend the BioNet for RGB imagery, we introduce a NeFF module to extract 3D features from drone imagery. NeFF reconstructs 3D crop structures and extracts semantic features from 2D images, learning geometry, radiance, and surface semantics. BioNet then uses these 3D features for final biomass prediction, yielding a 7.9\% improvement over SOTA methods. This scalable, vision-based approach has the potential for large-scale agricultural applications. Our contributions include:
\begin{itemize}
    \item A novel 3D backbone network combining sparse 3D CNN and Transformer encoders, outperforming SOTA methods in crop biomass prediction accuracy.
    \item The development of an implicit neural network that generates 3D features from drone imagery, showing superior scalability and accuracy over other vision-based biomass measurement methods.
    \item Extensive experiments on public datasets, including comprehensive evaluation and ablation studies.
\end{itemize}

This paper is structured as follows: we review related work in~\cref{sec:related_work}, describe our approach in~\cref{sec:BPN} and~\cref{sec:NeFF}, and present experimental results in~\cref{sec:experiments}, concluding in~\cref{sec:conclusion}.

\section{Related work}
\label{sec:related_work}
\vspace*{-0.2cm}

\begin{figure*}[h!]
\centering
\includegraphics[width=2.0\columnwidth]{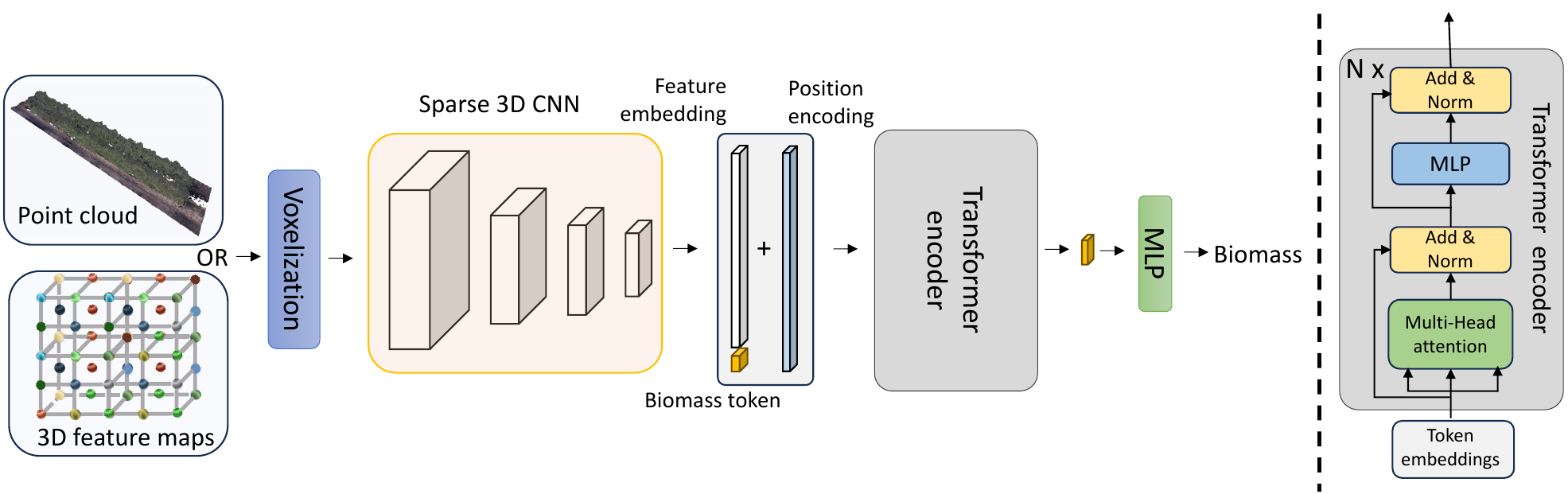}
\setlength{\belowcaptionskip}{-16pt}
\vspace*{-0.4cm}
\caption{The framework of our BioNet. Point clouds or 3D feature maps are first voxelized, and a sparse 3D CNN backbone then extracts 3D local structure features and compresses the height dimension into one. The final 2D feature map is flattened into a feature embedding, to which a learnable biomass embedding token (yellow cuboid) is appended. This biomass token acts as a comprehensive 3D feature representation and connects to a two-layer MLP for final biomass prediction. The feature embedding is added to a learnable position embedding before the Transformer encoder. The rightmost panel shows the structure of the Transformer encoder as inspired by~\cite{vaswani2017attention}.}
\label{fig:bpn}
\end{figure*}

\noindent \textbf{Biomass prediction methods}: The use of canopy height as a proxy for biomass estimation has recently gained traction~\cite{loudermilk2009ground, Gebbers2011, tilly2014multitemporal, ye2023uav}. Saeys et al.\cite{saeys2009estimation} employed statistical models to estimate crop density using LiDAR data. Tilly et al.\cite{tilly2014multitemporal} combined 3D and spectral data to create bivariate regression models~\cite{tilly2015fusion}. Li et al.\cite{li2015airborne} utilized airborne LiDAR and structural equation modeling to assess plant height and leaf area index. However, canopy height alone is insufficient for accurate biomass prediction when height variation is minimal. Other factors, such as point cloud volume and 3D indices, correlate strongly with measured biomass\cite{jimenez2018high, walter2019estimating, sun2018field, caballer2021prediction, loudermilk2009ground}. Point density, which captures the 3D structure of crops, has been explored through various methods~\cite{sun2018field, greaves2015estimating, polo2009tractor, walter2019estimating, Alonzo_2020}. Notably, Jimenez-Berni et al.~\cite{jimenez2018high} introduced a voxel-based method (3DVI), dividing point clouds into voxels to calculate non-empty voxel ratios. This method is considered a ‘gold standard’ for non-learning-based biomass estimation approaches~\cite{walter2019estimating, ten2019biomass}. Recent advancements in deep neural networks offer promising improvements in biomass prediction. Oehmcke et al.\cite{oehmcke2022deep} explored various architectures for point cloud regression, demonstrating substantial accuracy gains. Pan et al.\cite{pan2022biomass} developed a deep-learning model for point clouds, incorporating completion and attention-based fusion modules. In this work, we propose a novel biomass prediction network for both image and point cloud modalities.

\noindent \textbf{Learning 3D features from multi-view images}: The widespread availability, affordability, and portability of RGB cameras make multi-view image-based methods attractive for 3D tasks like 3D reconstruction\cite{Barron2022MipNeRF360, chen2024dehazenerf, yao2024neural, yariv2021volume, remondino2023nerfsurvey} and object detection\cite{philion2020lift, li2020object, huang2021bevdet, li2020real}. Extracting high-quality 3D features is crucial for these tasks, often achieved by projecting 3D grids onto 2D feature maps using camera matrices~\cite{roddick2018orthographic} or attention mechanisms~\cite{huang2021bevdet}. These explicit 3D geometrical representations, however, are memory-intensive and result in lower-resolution grids. An alternative approach lifts 2D features into 3D space by predicting depth distributions~\cite{philion2020lift}, but this requires dense point cloud supervision. In contrast, neural implicit representations offer a powerful, memory-efficient alternative for various 3D tasks. These representations have been widely applied in 3D reconstruction\cite{NeuS_3D_reconstru, yariv2021volume, rabby2023beyondpixels, Roessle2022DenseDepthPriors}, novel view synthesis\cite{mildenhall2021nerf, Turki2022Mega-NeRF, jia2024drone}, and feature extraction\cite{tschernezki2022neural, ye2023featurenerf, li2019detection, kobayashi2022decomposing}. Neural radiance fields (NeRF)\cite{mildenhall2021nerf, cai2024nerf, zhu2023DeepReviewNeRF} use implicit geometry and radiance fields to synthesize high-quality images from multi-view data. The Implicit Differentiable Renderer (IDR)\cite{yariv2020multiview, kellnhofer2021neural} leverages Signed Distance Functions (SDFs) to separate color and geometry fields, enhancing 3D surface representation. Neural implicit surfaces (NeuS)\cite{NeuS_3D_reconstru} and Volume SDF\cite{yariv2021volume} extend this with volumetric rendering to optimize 3D geometry. These methods outperform Structure from Motion (SfM)\cite{schonberger2016structure}. Semantic extensions, such as Semantic-NeRF\cite{zhi2021place}, further integrate multi-view semantic fusion with differentiable rendering. NeRF has been adapted to continuous feature fields by distilling 2D features from foundation models for scene editing\cite{kobayashi2022decomposing}, object retrieval\cite{tschernezki2022neural}, and semantic segmentation\cite{ye2023featurenerf}. For our application, accurate 3D structural and semantic feature extraction is crucial. Neural implicit representations are well-suited for 3D reconstruction and feature extraction, making them ideal for our biomass estimation tasks.

\vspace*{-0.2cm}
\section{Biomass prediction network}\label{sec:BPN}
\vspace*{-0.2cm}

Given a 3D input which can be either point cloud or 3D feature maps, ${\boldsymbol{P}} \in \mathbb{R}^{n \times (3+f)}$, where $n$ denotes the number of points, 3 represents its location dimension, and $f$ is the point feature channel number, (e.g., $f=3$ for colorized point clouds), Biomass prediction network is to take ${\boldsymbol{P}}$ as input and predict the biomass. Since the 3D structure of crops is inherently correlated with their biomass, we employ the sparse 3D CNN to extract 3D local features from ${\boldsymbol{P}}$. Moreover, sparse 3D CNNs are chosen for their efficiency in terms of less learning parameters and translation invariance, outperforming MLP and Transformer. We also exploit the power of Transformers~\cite{vaswani2017attention, dosovitskiy2020image, chen2021crossvit} to capture global features for robust biomass prediction. The hybrid architecture of our BioNet is shown in\cref{fig:bpn}.

\subsection{3D feature extraction}
The 3D input $\boldsymbol{P}$ typically contains 3D geometrical information but in a sparse representation, so we design a backbone network (denoted as $\mathcal{H}(.)$), mainly consisting of sparse 3D CNN layers, to extract 3D geometric features $\boldsymbol{f}^{3D}$, as shown in~\cref{equ:3dcnn}. Sparse 3D CNN takes advantage of spatial sparsity in point cloud data~\cite{s18103337, li2019three}, which can significantly reduce memory consumption and processing costs. The sparse 3D CNN expects its input, $\boldsymbol{P}$, to be organized within a regular grid, but the input point cloud data is usually in an irregular format. We therefore first voxelize the $\boldsymbol{P}$ to obtain its voxel representation. Each resulting voxel may contain several points, and we average all points inside a given voxel to obtain its final feature. Our voxelized feature maps with high resolutions equate to $c \times l \times w \times h$ where $c = (3+f)$. In our application, $h$ (height) is the smallest dimension as crop height tends to be less than plot width and length. The 3D features are then processed through sparse 3D CNN blocks, using a $3 \times 3 \times 3$ convolutional kernel and submanifold CNN~\cite{graham2017submanifold} until the height dimension collapses, resulting in a 2D feature map $\boldsymbol{f}^{3D} \in \mathbb{R}^{c^t \times l^t \times w^t}$. The full architecture and voxelization size are detailed in the supplementary material.
\vspace*{-0.2cm}
\begin{equation}
    \boldsymbol{f}^{3D} = \mathcal{H}(\boldsymbol{P})
    \label{equ:3dcnn}
\end{equation}

\subsection{Transformer for biomass prediction}
Transformer architecture has proven highly effective in various computer vision tasks \cite{carion2020end, dosovitskiy2020image, misra2021end, vaswani2017attention} due to its strong capability in feature extraction, making it an ideal choice for biomass prediction. After extracting the feature map $\boldsymbol{f}^{3D}$ from 3D sparse CNN, we flatten it into a 1-dimensional token sequence $\boldsymbol{f}^{3D}_{t} \in \mathbb{R}^{c^t \times (l^t\times w^t)}$ as the input for the Transformer. Similar to \texttt{Class} token in ViT \cite{dosovitskiy2020image}, we introduce a learnable embedding, called \texttt{Biomass} token, to the feature embedding $\boldsymbol{f}^{3D}_{t}$. The \texttt{Biomass} token serves as the 3D feature representation for the entire scene and interacts with overall spatial features. It can be treated as an agent extracting or summarizing all feature embeddings. Position embeddings are also added to each feature embedding to retain positional information. We used standard learnable 1D position embedding \cite{vaswani2017attention}. The final sequence of embedding vectors is passed through Transformer encoders. A biomass prediction head with two-layers MLP is connected with the \texttt{Biomass} token to predict the final biomass value $\hat{m}$. The entire Transformer prediction module plus two-layers MLP is represented as $\mathcal{F}$(.), and $\hat{m}$ is as follows:
\vspace*{-0.2cm}
\begin{equation}
    \hat{m} = \mathcal{F}(\boldsymbol{f}^{3D}_{t})
    \label{equ:trans}
\end{equation}

\begin{figure}[htb]
    \centering
    \includegraphics[width=0.8\columnwidth]{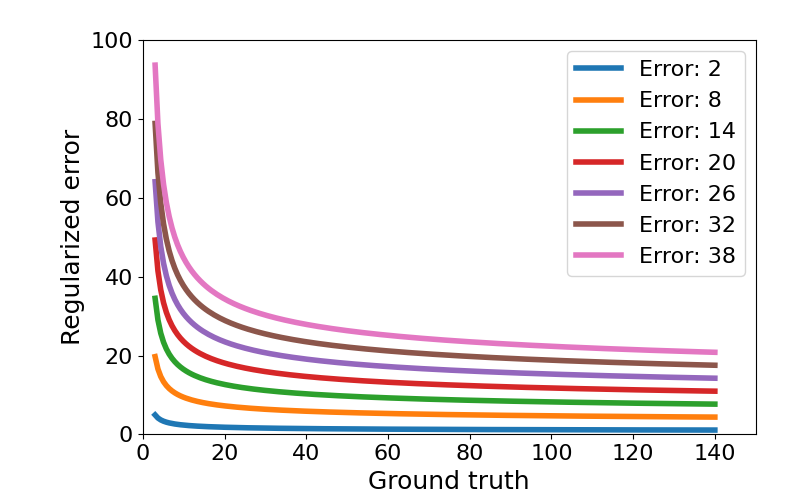}
    \setlength{\belowcaptionskip}{-12pt}
    \vspace*{-0.4cm}
    \caption{Graph showing the biomass prediction error against the regularized error for different ground truth values.}
    \label{fig:loss}
\end{figure}

Given the ground truth biomass $m_i$, we calculate the final biomass prediction loss as the following equation:
\begin{equation}
\begin{aligned}
\mathcal{L}_{b} =  \sum_{i=1}^{B} \boldsymbol\ell1\left( \frac{\hat{m}_i - m_i}{\log(m_i)} \right)
\label{equ:bio_pred}
\end{aligned}  
\end{equation}
\noindent where $\boldsymbol\ell1$ represents smooth 1-norm loss function~\cite{smoothl1} and $B$ is the batch size. To address the issue of the large varying ranges of biomass ground truth throughout the season, we regularize the error between prediction and ground truth using $\log(m_i)$, resulting in a loss function $\mathcal{L}_{b}$ to is more sensitive to small biomass number. Thus, prediction errors are penalized less as biomass increases. The relationship between error and regularized error is illustrated in~\cref{fig:loss}.
\vspace*{-0.3cm}

\begin{figure*}[htb]
    \centering
    \includegraphics[width=2.0\columnwidth]{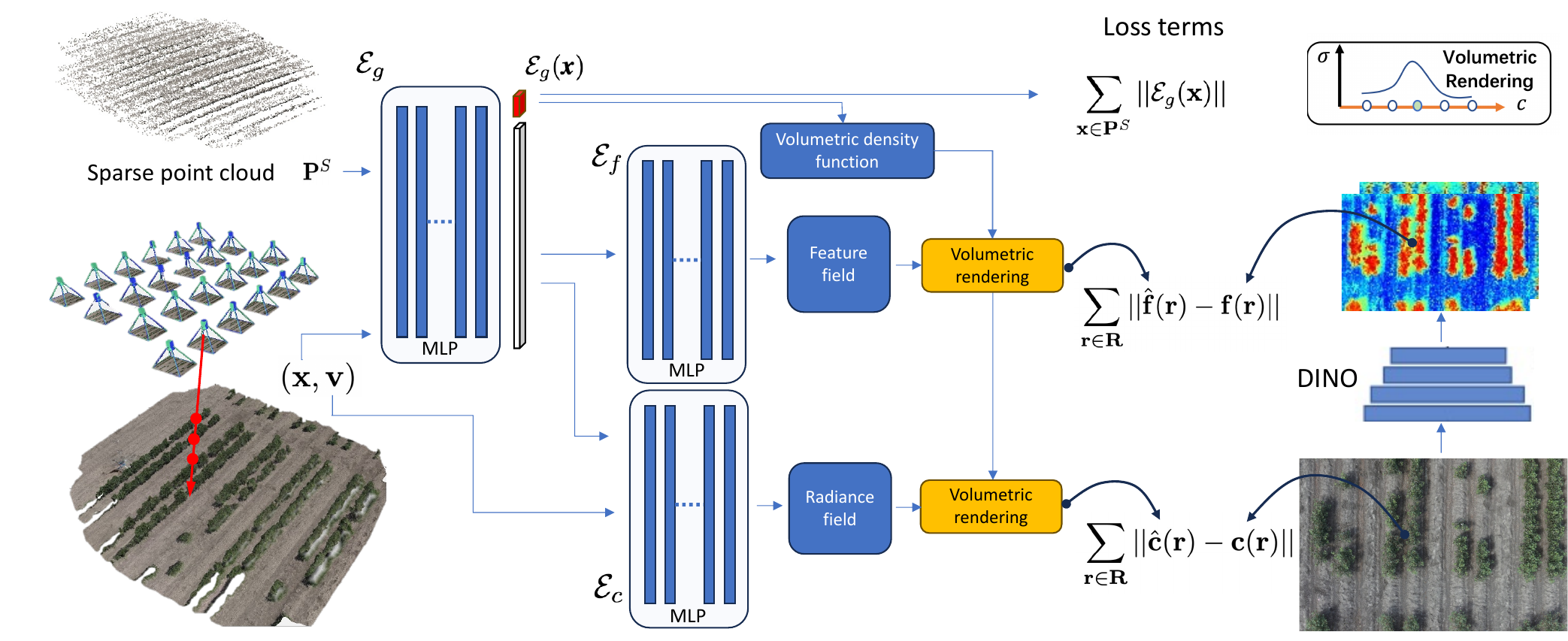}
    \setlength{\belowcaptionskip}{-16pt}
    \vspace*{-0.35cm}
    \caption{The framework of our NeFF module. The neural geometry field $\mathcal{E}_{g}$ takes as input the sampled points $\mathbf{x}$ along camera ray $\mathbf{r}(t)$ (red arrow) and sparse points $\textbf{P}^{S}$ from SfM. The red cuboid ($\mathcal{E}_{g}(x)$) represents SDF output for points belonging to $\textbf{P}^{S}$. The neural feature field $\mathcal{E}_{f}$ predicts the 3D semantic features for points $\mathbf{x}$, and the radiance field $\mathcal{E}_{c}$ will predict the colors of sampled 3D points $\mathbf{x}$ with viewing direction $\mathbf{v}$. The novel-view image and its feature maps are rendered through the volumetric rendering function in NeuS~\cite{NeuS_3D_reconstru}. 3D feature maps extracted from feature field $\mathcal{E}_{f}$ are used for input to the BioNet, shown in~\cref{fig:bpn}.}
    \label{fig:NeFF}
\end{figure*}

\section{Neural feature field}
\label{sec:NeFF}

The vision-based biomass prediction approach is particularly appealing due to the widespread availability of RGB cameras. To make the BioNet applicable to the vision system, we introduce a neural feature field (as shown in~\cref{fig:NeFF}) to extract 3D features from drone images. That is, given drone imagery of a static scene $(I_{1}, I_{2}, \ldots, I_{n})$, $I_i \in \mathbb{R}^{3 \times H \times W}$, neural feature field distills 2D features from foundation models into a 3D representation suitable for biomass prediction. These 3D features should not only encode generic semantic content, which aids in distinguishing different plant parts and in isolating them from the ground but also capture the 3D spatial plant structure.

\subsection{Neural implicit scene representation}
A scene’s representation consists of two main aspects: geometry and appearance. Similar to NeuS \cite{NeuS_3D_reconstru}, we represent scene geometry using a neural implicit surface with SDF, due to its enhanced 3D reconstruction with a better decoupling between the geometry and appearance information \cite{yariv2020multiview, NeuS_3D_reconstru, yariv2021volume}. Given a 3D location $\textit{\textbf{x}} = (x,y,z) \in \mathbb{R}^{3}$ in the specified 3D volume, the geometry mapping function, i.e. SDF, is defined as $\mathcal{E}_{g}: \textit{\textbf{x}} \longrightarrow \textit{s} $, and $s \in \mathbb{R}$ is the signed distance of \textbf{x} to the object's surface, approximated by a multi-layer perception (MLP). The surface $\textit{S}$ of the object is depicted by the zero-level set of its SDF as follows:
\vspace*{-0.2cm}
\begin{equation}
\begin{aligned}
 \textit{S} = \{ \textbf{\textit{x}} \in \mathbb{R}^{3} | \mathcal{E}_{g}(\textbf{\textit{x}}) = 0 \}
\label{equ:sdf}
\end{aligned}
\end{equation}
The scene appearance, i.e. radiance field, is represented by another independent MLP, and the color mapping function is $\mathcal{E}_{c}: (\mathbf{x}, \mathbf{v}) \longrightarrow \mathbf{c}$. Since the appearance is view-dependent, the viewing direction $\mathbf{v} \in \mathbb{R}^{2}$ is included in the input. The output $\mathbf{c} \in \mathbb{R}^{3}$ (i.e RGB) represents the color emitted from point $\mathbf{x}$ along the viewing direction $\mathbf{v}$. After getting both the geometry and appearance, volume rendering is applied to generate image colors. Given a camera ray $\mathbf{r}(t) = \mathbf{o} + t\mathbf{v}$, where $\mathbf{o}$ is the camera origin, the expected color of this ray is the integrated color along the ray with near ($t_{\text{n}}$) and far ($t_{\text{f}}$) bounds. This integration is computed as:
\vspace*{-0.2cm}
\begin{equation}
\begin{aligned}
\mathbf{\hat{c}(r)} =  \int_{t_{\text{n}}}^{t_{\text{f}}} w(t)\mathbf{c}(\mathbf{x}(t), \mathbf{v})dt  =   \int_{t_{\text{n}}}^{t_{\text{f}}} w(t)\mathcal{E}_{c}(\mathbf{x}(t), \mathbf{v})dt
\label{equ:volume}
\end{aligned}
\end{equation}
\noindent where $\mathbf{x}(t)$ is sampled 3D point along ray $\mathbf{r}$ at $t$ positon and $w(t)$ is its integration weight. The $w(t)$ can be derived by learning an approximated function from SDF $\mathcal{E}_{g}(t)$, the details of which can refer to paper~\cite{NeuS_3D_reconstru}. With the geometry and appearance information, we can synthesize an image from any novel view via volume rendering.

\subsection{Semantic feature representation}
\label{sec:feature_field_rep}
To generate a continuous 3D semantic feature volume, we use a neural feature field to distill the knowledge from the 2D foundation model, DINO \cite{caron2021emerging}, via neural rendering. The 2D foundation model, trained on the large image dataset, generally has a strong transferability~\cite{oquab2023dinov2}, and the distilled knowledge can help to overcome the shortage of labeled data in the domain dataset. The extracted features from the foundation model for each image are $(F_{1}, F_{2}, \ldots, F_{n}), F_i \in \mathbb{R}^{c_i \times h_i \times w_i}$, where $c_i$ is the feature dimension. The semantic feature field is view-independent, and its mapping function can be defined as $\mathcal{E}_{f}: \mathbf{x} \longrightarrow \mathbf{f}$, where $\mathbf{f} \in \mathbb{R}^{c_i}$. Similar to learning the radiance field via rendering color, We learn the feature field $\mathcal{E}_{f}$ by the volumetric rendering of the 2D feature map $F_i$. Therefore, the expected 2D feature for a camera ray, $\mathbf{r}(t) = \mathbf{o} + t\mathbf{v}$, can be represented as~\cref{equ:feature}, in which the scene geometry $w(t)$ is the same to~\ref{equ:volume}.
\vspace*{-0.2cm}
\begin{equation}
\begin{aligned}
\mathbf{\hat{f}(r)}  = \int_{t_{\text{n}}}^{t_{\text{f}}} w(t)\mathbf{f}(\mathbf{x}(t))dt =   \int_{t_{\text{n}}}^{t_{\text{f}}} w(t)\mathcal{E}_{f}(\mathbf{x}(t))dt
\label{equ:feature}
\end{aligned}
\end{equation}
Once the semantic feature field $\mathcal{E}_{f}$ is well learned, we can extract the 3D semantic feature $F_{3D}$ by inputting the surface points $\textit{S}$ (i.e,~
\cref{equ:sdf}), as follows
\vspace*{-0.1cm}
\begin{equation}
    F_{3D} = \{\mathcal{E}_{f}(\textbf{\textit{x}}) | \textbf{\textit{x}} \in \mathbb{R}^{3}, \mathcal{E}_{g}(\textbf{\textit{x}})=0\}
    \label{equ:3DF_EXT}
\end{equation}

\subsection{Loss function}
If three neural representation functions (i.e. $\mathcal{E}_g$, $\mathcal{E}_{c}$, and $\mathcal{E}_{f}$) are well learned, we should be able to render both image and feature map with high-fidelity. Therefore, given a camera ray $\mathbf{r}$, we can construct the color loss function to minimize the difference between the rendered image $\hat{\mathbf{c}}(\mathbf{r})$ and ground truth one $\mathbf{c}(\mathbf{r})$, as following equation.
\vspace*{-0.2cm}
\begin{equation}
\begin{aligned}
 \mathcal{L}_{c} =  \sum_{\mathbf{r} \in \mathbf{R}} || \hat{\mathbf{c}}(\mathbf{r}) - \mathbf{c}(\mathbf{r}) ||
\label{equ:loss_color}
\end{aligned}
\end{equation}
\noindent where $\mathbf{R}$ is the sampled rays for training. Similarly, the feature loss function can be built to optimize function $\mathcal{E}_{f}$.
\vspace*{-0.1cm}
\begin{equation}
\begin{aligned}
 \mathcal{L}_{f} =  \sum_{\mathbf{r} \in \mathbf{R}} || \hat{\mathbf{f}}(\mathbf{r}) - \mathbf{f}(\mathbf{r}) ||
\label{equ:loss_feature}
\end{aligned}
\end{equation}
In practice, we find that these two losses are not strong enough to recover the geometry information, seeing~\cref{fig:depth}. This can be caused by two reasons, one is that plants are quite small and most area in the scene is quite boring and flat, other is that the $\mathcal{E}_{g}$ is implicitly learned with color and features with direct supervision. Given drone imagery, we run the SfM \cite{schonberger2016structure} to obtain the camera parameters, and a sparse point cloud is generated as a side-product. To better optimize the $\mathcal{E}_{g}$, we use this sparse point cloud to directly optimize $\mathcal{E}_{g}$. The constructed loss is as follows:
\vspace*{-0.2cm}
\begin{equation}
\begin{aligned}
 \mathcal{L}_{g} =  \sum_{\mathbf{x} \in \mathbf{P}^{S}} || \mathcal{E}_{g}(\mathbf{x})||
\label{equ:loss_geometrical}
\end{aligned}
\end{equation}
\noindent where $\mathbf{P}^{S} \in \mathbb{R}^{k \times 3}$ is denoted as sparse point clouds from SfM and $k$ is the number of points. The final loss function for optimizing the NeFF is as follows, and $\alpha$ is a hyperparameter (we set 0.002 in our experiment).
\begin{equation}
\begin{aligned}
 \mathcal{L}_{NeFF} = \mathcal{L}_{c} + \mathcal{L}_{g} + \alpha \mathcal{L}_{f} 
\label{equ:loss_totall}
\end{aligned}
\end{equation}

\vspace*{-0.2cm}

\begin{table*}[htb]
    \centering
    \scalebox{0.95}{
    \begin{tabular}{c c | c  c  c c c l} 
    \specialrule{.12em}{.12em}{.12em}
    \multicolumn{2}{c|}{\backslashbox{Metric}{Methods}} & PointNet\cite{qi2017pointnet} & PoinNet++\cite{qi2017pointnet++} & DGCNN\cite{wang2019dynamic} & PC-BioNet \cite{pan2022biomass} & 3DVI \cite{jimenez2018high} &  Ours \tiny{(RI)}\\
    \specialrule{.12em}{.12em}{.12em}
     \multirow{3}{*}{\makecell{Imagery\\ w \small{NeFF}} } & MAE    & 175.9  & 165.8  &\cellcolor{third} 153.1  & \cellcolor{second}150.7  & 193.4 & \cellcolor{best}138.8  \scriptsize{(7.9\%)}  \\ 
      & MARE   & 0.316  & 0.294  & \cellcolor{third}0.280  & \cellcolor{second} 0.278  & 0.398 &  \cellcolor{best}0.231 \scriptsize{(16.9\%)}\\ 
      & RSME        & 265.8  & 246.1  &\cellcolor{third}  229.4  & \cellcolor{second}223.9  & 302.5 & \cellcolor{best}197.2 \scriptsize{(11.9\%)} \\  \specialrule{.12em}{.12em}{.12em}

    \multirow{3}{*}{\makecell{Imagery\\ w \small{SfM}} } & MAE    & 177.2  & 166.2  &\cellcolor{third} 156.7  & \cellcolor{second}153.8  & 206.2 & \cellcolor{best}145.3  \scriptsize{( 5.5\%)}  \\ 
      & MARE   & 0.320  & 0.311  & \cellcolor{second}0.283  & \cellcolor{third} 0.292  & 0.428 &  \cellcolor{best}0.261 \scriptsize{(7.8\%)}\\ 
      & RSME        & 273.3  & 252.8  &\cellcolor{third}  233.6  & \cellcolor{second}228.4  & 314.2 & \cellcolor{best}213.5 \scriptsize{( 6.5\%)} \\  \specialrule{.12em}{.12em}{.12em}  
     \multirow{3}{*}{\makecell{Point \\ clouds}} & MAE    & 172.4 & 164.8 & 133.1 &  \cellcolor{second}128.8  &  \cellcolor{third}  130.3  & \cellcolor{best}120.9  \scriptsize{(6.1\%)}  \\ 
      & MARE   & 0.253 & 0.229 & 0.236 &  \cellcolor{second}0.201 & \cellcolor{third}  0.225 & \cellcolor{best}0.174 \scriptsize{(13.4\%)}\\ 
      & RSME    & 256.3 & 215.6 & \cellcolor{third}  203.6 &  \cellcolor{second}194.5 & 204.8 & \cellcolor{best}180.1 \scriptsize{(7.4\%)}\\  \specialrule{.12em}{.12em}{.12em}
    \end{tabular}}
    \setlength{\belowcaptionskip}{-12pt}
    \vspace*{-0.2cm}
    \caption{Quantitative performance comparisons of our method (in the rightmost column) with baselines for point clouds and imagery on MMCBE. The color of each cell indicates the \colorbox{best}{best}, \colorbox{second}{second}, and \colorbox{third}{third}s score. For Imagery w SfM, the input for all methods is pseudo points obtained with SfM \cite{schonberger2016structure}. The Imagery w NeFF represents that input for baselines is extracted from the geometrical field $\mathcal{E}_{g}$ in NeFF. The unit for MAE and RSME is the gram. Some scores of baselines were reported in~\cite{li2024mmcbe}.}
    \label{tab:exp_results}
\end{table*}

\section{Experiments}
\label{sec:experiments}

\subsection{Experimental Setup}
\label{sec:exp}

\noindent \textbf{Dataset}: Two public datasets, a wheat dataset~\cite{pan2022biomass} and MMCBE~\cite{li2024mmcbe}, were used to validate our method. The wheat dataset comprises 306 plots with point clouds only, collected at two timepoints by a mobile robot, focusing on wheat crops. It was divided into 204 training plots and 102 testing plots. Conversely, MMCBE consists of 216 plots, capturing Camera and LiDAR data on cotton crops across 9 timepoints, and was partitioned into 152 training and 64 testing sets. Notably, these datasets cover different crop species and structures, with wheat being a determinate monocotyledon grass species and cotton an indeterminate dicotyledon shrub species.

\noindent \textbf{Training and Augmentation}: To prevent overfitting, we used two data augmentation methods: rotation and motion. Rotation involves random angle adjustments ($\theta$, $\alpha$, $\phi$) around x, y, z-axes within ranges ($\left[-\frac{\pi}{18}, \frac{\pi}{18}\right]$, $\left[-\frac{\pi}{18}, \frac{\pi}{18}\right]$, $\left[-\frac{\pi}{12}, \frac{\pi}{12}\right]$). Motion introduces Gaussian-distributed translation offsets ($\delta x, \delta y, \delta z$) with zero mean and unit standard deviation. More training details are explained in supplementary materials.

\noindent \textbf{Baselines}: Our method was compared to that of existing SOTA methods, namely 3DVI~\cite{jimenez2018high, walter2019estimating}, BioNet (Referred to hereafter as PC-BioNet for Point Cloud-BioNet)~\cite{pan2022biomass}, PointNet~\cite{qi2017pointnet}, PointNet++~\cite{qi2017pointnet++}, and DGCNN~\cite{wang2019dynamic}. Given that they all take point clouds as input, we used SfM~\cite{schonberger2016structure} to reconstruct point clouds from our RGB data and trained all methods on our dataset.

\noindent \textbf{Evaluation metrics}: We used similar metrics to evaluate biomass prediction errors as in \cite{pan2022biomass}, namely mean absolute error \textit{(MAE)}, mean absolute relative error (\textit{MARE}), and root mean square error \textit{(RMSE}). The lower these errors, the better the prediction accuracy. RI is used to indicate the relative improvement when comparing two different methods, defined in supplementary materials. 

\begin{table*}[h!]
    \centering
    \scalebox{0.9}{
    \begin{tabular}{c  c | c  c  c c c l} \specialrule{.12em}{.12em}{.12em}
    \multicolumn{2}{c|}{\backslashbox{Metric}{Methods}} & PointNet\cite{qi2017pointnet} & PoinNet++\cite{qi2017pointnet++} & DGCNN\cite{wang2019dynamic} & PC-BioNet \cite{pan2022biomass} & 3DVI \cite{jimenez2018high} & OURS \scriptsize{(RI)} \\  \specialrule{.12em}{.12em}{.12em}
     \multirow{3}{*}{\makecell{Overall}} & MAE  & 139.6 & 142.8 & 129.7 & \cellcolor{second}71.2 & \cellcolor{third} 115.2 & \cellcolor{best}51.8 \scriptsize{(27.3\%)} \\ 
                                    & MARE & 0.274 & 0.275 & 0.254 & \cellcolor{second}0.121 & \cellcolor{third} 0.190 & \cellcolor{best} 0.097 \scriptsize{(19.8\%)}\\ 
                                            & RSME  & 189.8 & 188.9 & 161.6 & \cellcolor{second}99.3 & \cellcolor{third} 151.8 & \cellcolor{best} 69.7 \scriptsize{(29.8\%)}\\  \specialrule{.12em}{.12em}{.12em}
     \multirow{3}{*}{\centering \makecell{Early \\ Stage}}& MAE & 92.6 &120.6 & 83.1 & \cellcolor{second}43.1 &\cellcolor{third} 60.5 & \cellcolor{best}35.1 \scriptsize{(18.7\%)}\\ 
                                                        & MARE & 0.326 &0.411 & 0.334 & \cellcolor{second}0.135 &\cellcolor{third} 0.211 & \cellcolor{best}0.121 \scriptsize{(10.4\%)} \\ 
                                                        &  RSME & 108.5 &159.2 &98.7 & \cellcolor{second}57.7 &\cellcolor{third} 69.8 &  \cellcolor{best}42.3 \scriptsize{(26.7\%)} \\  \specialrule{.12em}{.12em}{.12em}
     \multirow{3}{*}{\centering  \makecell{Flowering \\ Stage}}  & MAE &188.5 &181.7 &\cellcolor{third} 170.1 & \cellcolor{second}100.5 &172.0 & \cellcolor{best}62.4 \scriptsize{(37.9\%)} \\ 
                                                      &  MARE &0.219 &0.179 &\cellcolor{third} 0.164 & \cellcolor{second}0.106 &0.168 & \cellcolor{best}0.066 \scriptsize{(27.3\%)}\\ 
                                                      &  RSME &247.5 &231.0 &207.7 & \cellcolor{second}129.1 &\cellcolor{third} 204.8 & \cellcolor{best}84.3 \scriptsize{(34.7\%)}\\  \specialrule{.12em}{.12em}{.12em}
    \end{tabular}}
    \setlength{\belowcaptionskip}{-16pt}
    \vspace*{-0.2cm}
    \caption{Quantitative comparison with baselines for point cloud modality on wheat biomass dataset~\cite{pan2022biomass}. The model, trained on the overall dataset, is tested at different stages. Our method outperforms all baselines, especially in the Flowering Stage. Scores of baselines were reported in~\cite{pan2022biomass}.  } 
    \label{tab:resulst_on_wheat}
\end{table*}


\subsection{Experiments of BioNet}
\noindent \textbf{Comparison with other methods:} Two public datasets and two modalities (i.e. images and point clouds) were used to evaluate our method. The performance for both point cloud and image modality were evaluated on MMCBE~\cite{li2024mmcbe}, as shown in~\cref{tab:exp_results}. Our method consistently demonstrates better prediction accuracy compared to the baselines, the RI in MARE against the SOTA are 16.9\% and 13.4\% for image and point cloud modality respectively. As for image modality, the model's input can be derived from either SfM~\cite{schonberger2016structure} or NeFF. We found that the pseudo points from NeFF (i.e. Imagery w NeFF) result in better biomass prediction than those from SfM (i.e. Imagery w SfM), suggesting that NeFF likely produces more accurate 3D reconstructions than SfM. When applying the same approach, the point cloud modality exhibits higher prediction accuracy than the image modality, such as 0.174 (MARE) for the point cloud against 0.231 (MARE) for the image. This distinction is likely attributed to the superior quality and more accurate geometrical structure of point clouds in contrast to the pseudo points generated from multi-view images.

Our approach was also tested on the wheat dataset~\cite{pan2022biomass} which only provides point cloud modality. The experimental results are shown in~\cref{tab:resulst_on_wheat}. The model was trained on the overall dataset and then evaluated on different stages, using the same training\slash testing list as in~\cite{pan2022biomass}. Our RI in MARE is 19.8\%, 10.4\%, and 27.3\% for the overall, early, and late stages, respectively. These results indicate that our method generates more accurate biomass predictions in the late stage when crop structures become mature, clear, and distinguishable. It is also interesting to note that, despite using the same method and modality (i.e., point cloud), the performance significantly varies between the two datasets. The MAE and RSME on the wheat dataset are considerably lower than those on MMCBE. This discrepancy can be partially attributed to the differences in biomass scales between the datasets: the wheat dataset features plots of $1\times1~m^{2}$, whereas MMCBE’s plots are much larger at $1\times15~m^{2}$ and contain higher biomass values. Furthermore, the MARE, a more robust indicator against biomass scale variations, suggests that the same approach achieves superior results on the wheat dataset~\cite{pan2022biomass} compared to MMCBE~\cite{li2024mmcbe}. We visualized the difference in the supplementary material.


\begin{table}[h]
    \begin{adjustwidth}{-0.4cm}
    {}
    \scalebox{0.7}{
    \begin{tabular}{c c|c c c|c c c} \specialrule{.12em}{.12em}{.12em}
    \multicolumn{2}{c|}{\multirow{2}{*}{\backslashbox{Training}{Testing}}} & \multicolumn{3}{c}{Early stage} & \multicolumn{3}{|c}{Flowering stage} \\
    \multicolumn{2}{c|}{} & DGCNN & PC-BioNet  & Ours & DGCNN & PC-BioNet & Ours \\ \specialrule{.12em}{.12em}{.12em}
    \multirow{3}{*}{\centering \makecell{Early \\ Stage}} & MAE  & 101.8 & 50.5 & \cellcolor{best} 34.8  & 592.9 & \cellcolor{best} 121.9 & 130.8  \\
    & MARE  & 0.247 & 0.134 & \cellcolor{best} 0.124 & 0.658 & \cellcolor{best} 0.137 & 0.158 \\
    & RMSE  & 116.1 & 62.2 & \cellcolor{best} 43.3 & 613.6 & \cellcolor{best}226.5 & 242.6 \\ \specialrule{.12em}{.12em}{.12em}
    \multirow{3}{*}{\centering \makecell{Flowering \\ Stage}} & MAE  & 563.1 & 114.5 &\cellcolor{best} 62.1 & 133.3 & 111.1 & \cellcolor{best} 58.6 \\
    & MARE  & 1.476 & 0.295 & \cellcolor{best} 0.163 & 0.115& 0.099 & \cellcolor{best} 0.063 \\
    & RMSE  & 566.3 & 124.5 & \cellcolor{best} 74.8 & 150.2 & 145.7 & \cellcolor{best} 80.2 \\ \specialrule{.12em}{.12em}{.12em}
    \end{tabular}
    }
    \end{adjustwidth}
    \setlength{\belowcaptionskip}{-16pt}
    \vspace*{-0.2cm}
    \caption{Quantitative performance of adaptability on wheat biomass dataset across different growth stages.}
    \label{tab:generalizaton_on_wheat_biomass}
\end{table}

\begin{figure}[h]
    \centering
    \includegraphics[width=0.85\linewidth]{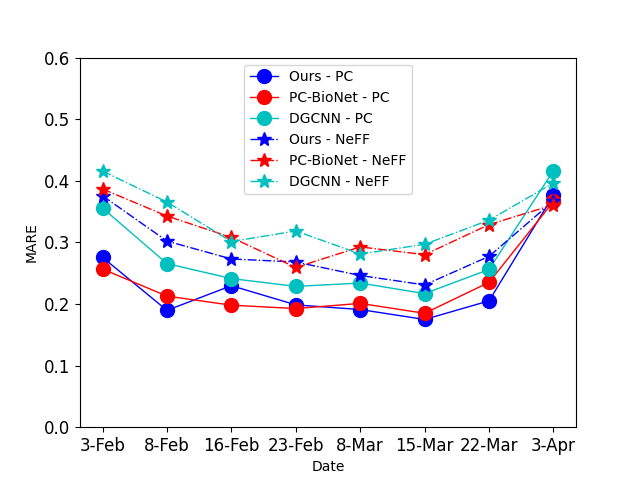}
    \setlength{\belowcaptionskip}{-12pt}
    \vspace*{-0.2cm}
    \caption{Adaptation capability on MMCBE. ``PC" is when the method takes point cloud as 3D input, while ``NeFF" indicates that 3D input is generated from the NeFF module.}
    \label{fig:generalization_our_dataset}
\end{figure}

\noindent \textbf{Adaptation capability:} As both biomass datasets are relatively small, the adaptation capability of models is important to their real-world applications. We assessed the adaptation capability on both datasets with the top 3 methods. For the MMCBE dataset, we used data from one timepoint as testing samples and trained the model on the data from the other timepoints, with the results shown in~\cref{fig:generalization_our_dataset}. The model can be adapted to data at middle timepoints after being exposed to both future and past data. However, its capability is still limited when adapted from data of past time points to future data, and vice versa. Nevertheless, our method exhibits a stronger adaptation than other methods. The testing results on the wheat biomass dataset can be found in~\cref{tab:generalizaton_on_wheat_biomass}. The model was trained with the Early Stage data and tested on both stages. In a separate experiment, the model was trained with the Flowering Stage data and tested on both stages. The results show that our model can adapt well from the flowering to the early stage, while PC-Bionet~\cite{pan2022biomass} has better adaptation from the early to flowering stage. A possible reason for this result is that the well-established 3D geometrical structure in the flowering stage could enhance the feature extraction and potentially increase the adaptation capability of our method.


\subsection{Experiments of NeFF module}
\vspace*{-0.2cm}
\begin{figure}[h]
    \centering
    \includegraphics[width=.8\columnwidth]{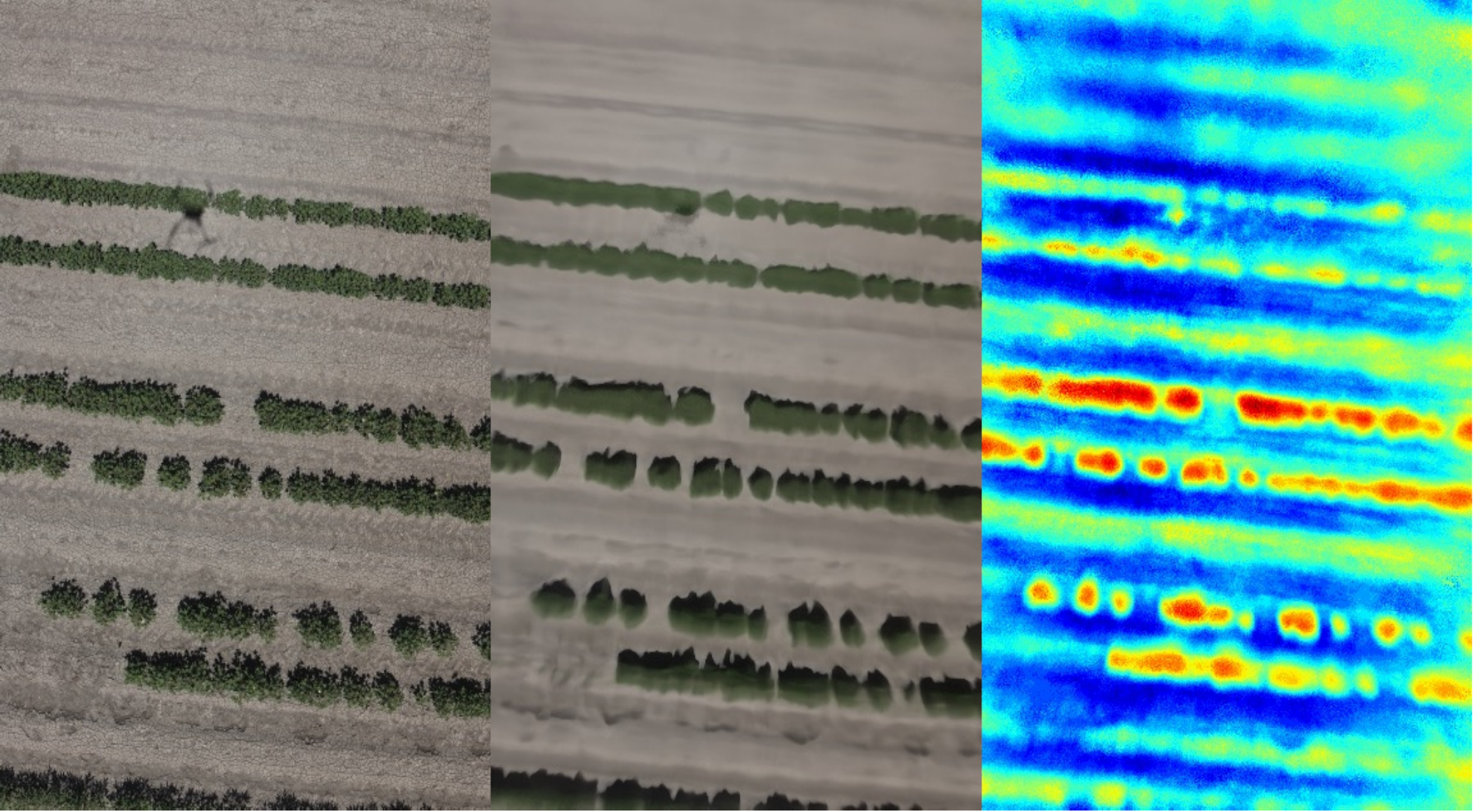}
    \setlength{\belowcaptionskip}{-16pt}
    \caption{Rendering results showing a raw image (left), a rendered image (middle), and a rendered feature map (right) with PCA compressing feature map into 3 channels.}
    \label{fig:rendering}
\end{figure}

\noindent \textbf{Rendering} We compared the rendering performance of our proposed method with that of NeuS~\cite{NeuS_3D_reconstru}, a SOTA method that has a similar image rendering module to ours. The comparison results can be found in~\cref{tab:PSNR}. The quality of our novel view synthesis was evaluated in terms of peak signal-to-noise ratio (PSNR) and structural similarity index (SSIM). We observed that the NeFF achieves a similar image rendering performance to NeuS despite including other additional tasks. This proves that the radiance field $\mathcal{E}_{c}$, feature field $\mathcal{E}_{f}$, and scene geometry modules $\mathcal{E}_{g}$ are well decoupled, with each part trying to learn its own representation. We also observed that the NeFF without $\mathcal{L}_{g}$ has a lower image rendering quality, the possible reason is that the scene geometry module $\mathcal{E}_{g}$ with implicit supervision is simultaneously optimized by both the radiance and feature fields, and the feature field $\mathcal{E}_{f}$ may hurt learning the radiance field $\mathcal{E}_{c}$ via the geometry module $\mathcal{E}_{g}$. The rendered image can be found in~\cref{fig:rendering}. 

\begin{table}[h!]
    \centering
    \scalebox{0.8}{
    \begin{tabular}{c|c|c|c|c|c|c} 
    \specialrule{.1em}{.1em}{.1em}
    \multicolumn{2}{c|}{ Scene ID} & \#1 & \#3 & \#5 & \#7 & \#9 \\ \specialrule{.1em}{.1em}{.1em}
    \multirow{3}{*}{\centering PSNR$\uparrow$} & NeuS & 25.41 & \textbf{23.31} & 21.92 & \textbf{21.6} & 22.54  \\\cline{2-7}
    & w/o NeFF & 24.19 & 22.76 &  21.21 &  20.32& 22.05 \\ \cline{2-7}
    & NeFF & \textbf{25.86} & 23.09 & \textbf{22.03} & 20.99  & \textbf{22.76} \\ \specialrule{.1em}{.1em}{.1em}
    \multirow{3}{*}{\centering SSIM$\uparrow$} & NeuS   & 0.90 & \textbf{0.89} & 0.82& \textbf{0.82}& 0.85 \\\cline{2-7}
                                               & w/o NeFF & 0.84 & 0.81 & 0.79& 0.73& 0.83\\ \cline{2-7}
                                               & NeFF    & \textbf{0.92} & 0.87 & \textbf{0.84} & 0.77 &\textbf{0.86} \\ \hline
    \end{tabular}}
    \setlength{\belowcaptionskip}{-10pt}
    \vspace*{-0.2cm}
    \caption{Quantitative comparison of our \textit{'NeFF'} method with NeuS on the task of novel-view image rendering on 1st, 3rd, 5th, 7th, and 9th timepoints. Rendering results of all image sets in a single timepoint were averaged. \textit{'w/o NeFF'} is our method without sparse point cloud supervision. Scores of NeuS were reported in~\cite{li2024mmcbe}.}
    \label{tab:PSNR}
\end{table}

\noindent \textbf{3D reconstruction} Due to the complexity of collecting ground truth in a real-world scenario, the MMCBE ground truth was limited and we consequently chose to qualitatively evaluate the 3D reconstruction, as shown in~\cref{fig:depth}. NeFF without sparse point cloud supervision (i.e. $\mathcal{L}_{g}$) leads to a rather flat and ambiguous 3D structure, whereas NeFF with $\mathcal{L}_{g}$ significantly improves the quality of 3D reconstructions, which can further enhance the quality of 3D feature representation. A 3D feature map $F_{3D}$, extracted from~\cref{equ:3DF_EXT}, is depicted in~\cref{fig:3d_feature_map}. The high-dimensional feature space was reduced into 3 channels with PCA for visualization and the visualized feature maps clearly illustrate that different parts of the crop can be easily distinguished using these distilled semantic features, underscoring the effectiveness of our extracted 3D feature maps.
\vspace*{-0.2cm}
\begin{table}[h!]
    \centering
    \scalebox{0.9}{
    \begin{tabular}{c|c|c|c|c} 
    \specialrule{.1em}{.1em}{.1em}
    \multicolumn{2}{c|}{ Input} & MAE  & MARE (\%)  & RSME  \\ \specialrule{.1em}{.1em}{.1em}
    \multirow{2}{*}{\centering NeFF w/o $\mathcal{L}_{g}$ } &xyz & 235.9 &	0.497 &	383.4  \\\cline{2-5}
                                 & feature  & 220.6 & 0.464 & 349.5 \\ \specialrule{.1em}{.1em}{.1em}
    \multirow{2}{*}{\centering NeFF w $\mathcal{L}_{g}$} & xyz  & 145.8	& 0.252	& 211.6 \\\cline{2-5}
                                 & feature  & \textbf{138.8} & \textbf{0.231} & \textbf{197.2} \\ \hline
    \end{tabular}}
    \setlength{\belowcaptionskip}{-16pt}
    \vspace*{-0.2cm}
    \caption{Ablation study on NeFF supervision and BioNet inputs. \textit{'NeFF w/o $\mathcal{L}_{g}$'} is our method without $\mathcal{L}_{g}$. \textit{'xyz'} and \textit{'feature'} represent BioNet inputs being either geometry coordinates or semantic features, respectively.}
    \label{tab:feat_depth_sup}
\end{table}


\begin{figure}[htb]
     \centering
     \includegraphics[width=0.9\columnwidth, height=4 cm]{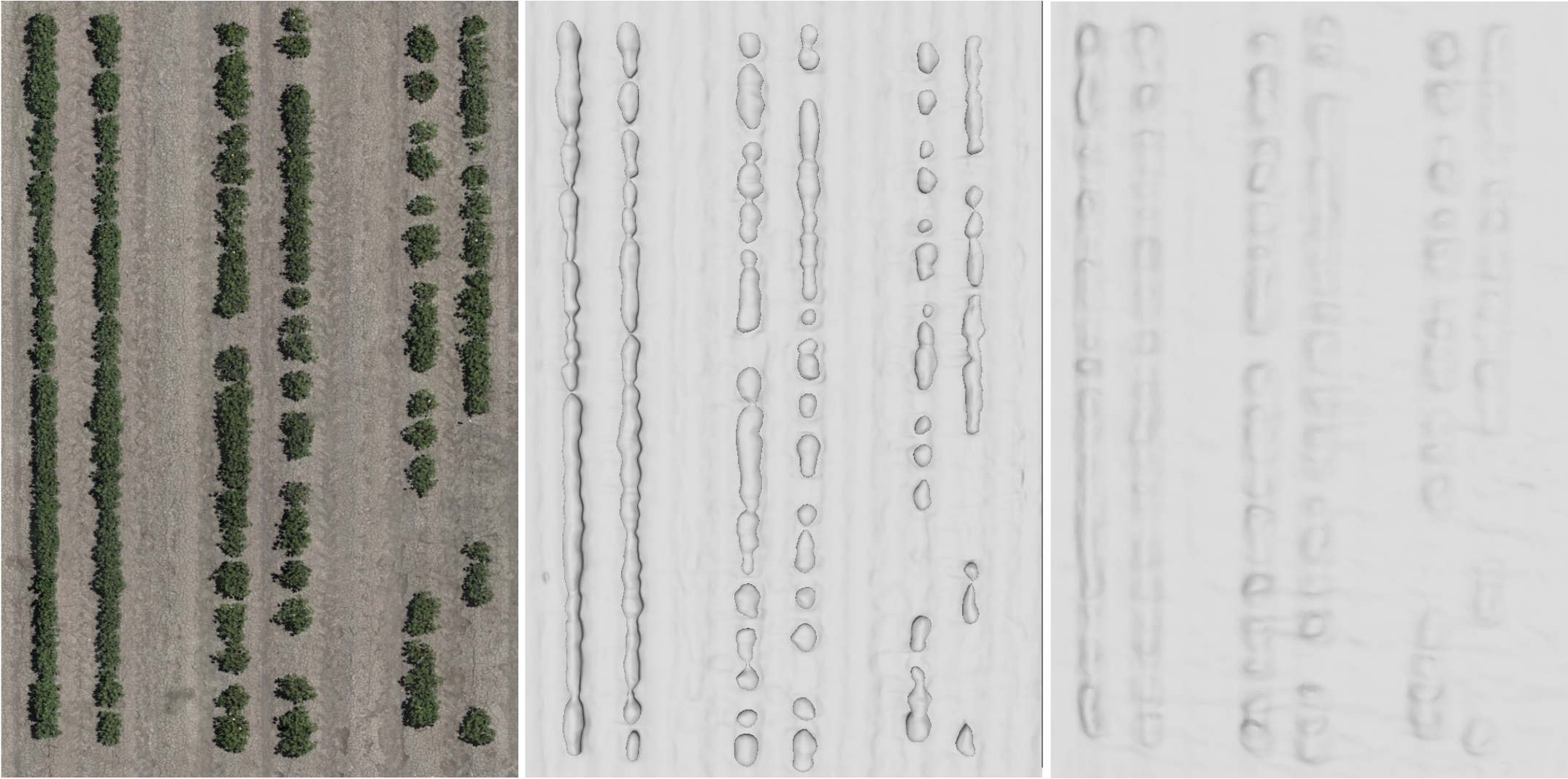}
     \setlength{\belowcaptionskip}{-12pt}
     \vspace*{-0.2cm}
    \caption{Visualization of 3D reconstruction and showing RGB image (left), 3D reconstructions from NeFF (middle) with $\mathcal{L}_{g}$, and reconstruction without $\mathcal{L}_{g}$ (right).}
    \label{fig:depth}
\end{figure}

\begin{figure}[htb]
     \centering
     \includegraphics[width=0.9\columnwidth, height=4cm]{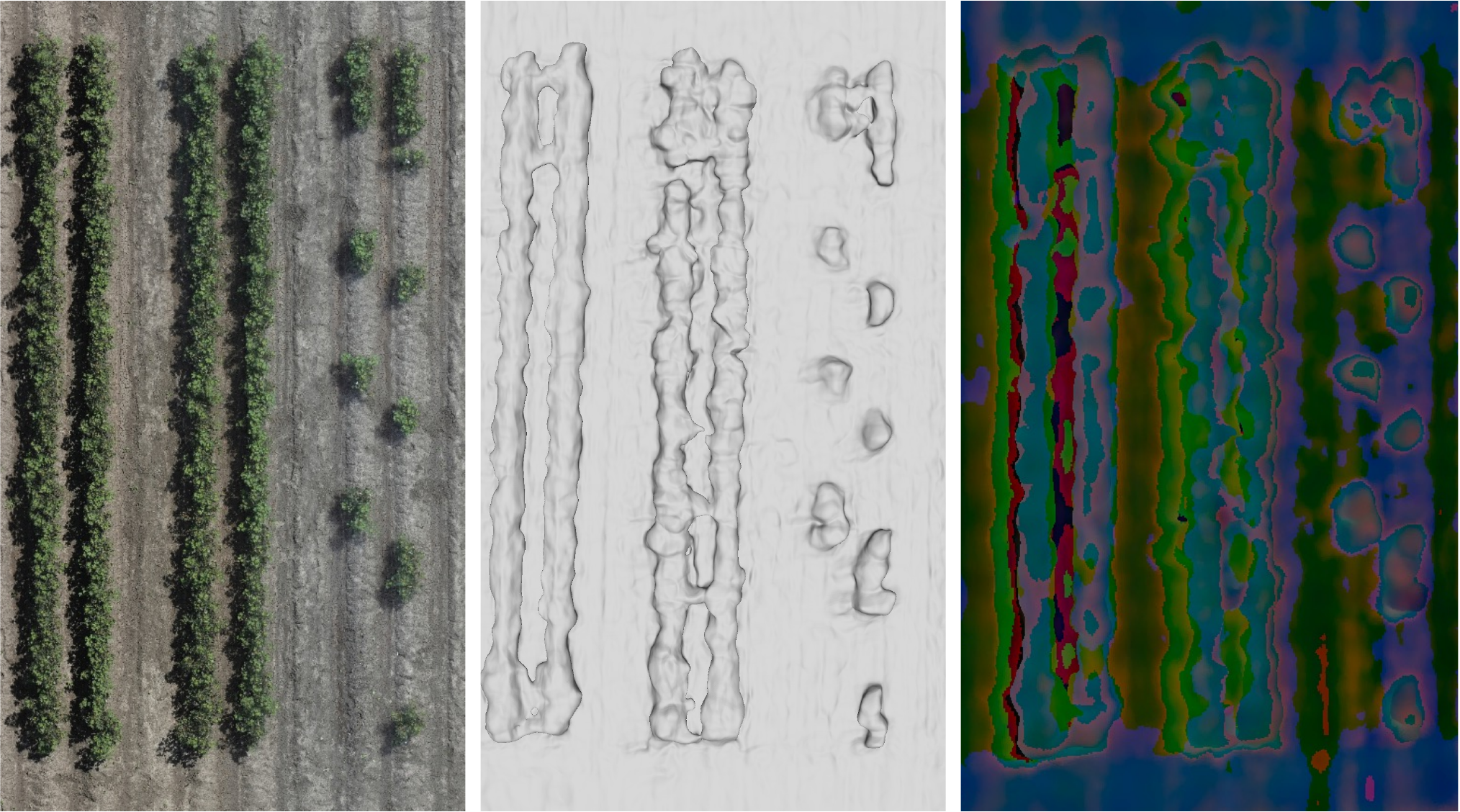}
     \setlength{\belowcaptionskip}{-12pt}
     \vspace*{-0.2cm}
    \caption{Visualization of 3D reconstruction and showing RGB images (left), 3D reconstructions from NeFF (middle), and 3D feature maps extracted from the feature field $\mathcal{E}_{f}$ (right)}
    \label{fig:3d_feature_map}
\end{figure}

\subsection{Ablation experiments on each module}
\begin{figure}[htb]
     \centering
     \includegraphics[width=0.8\columnwidth]{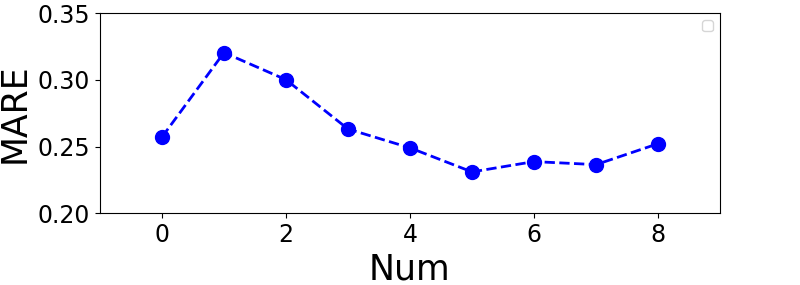}
      \setlength{\belowcaptionskip}{-12pt}
     \caption{Ablation study on the number of Transformer encoders in our BioNet module.}
     \vspace*{-0.2cm}
     \label{fig:transformer_encoder}
\end{figure}

\vspace*{-0.2cm}
Several ablation studies were performed over the MMCBE dataset. First, we tested whether the use of sparse point cloud supervision in NeFF could improve prediction accuracy. As shown in~\cref{tab:feat_depth_sup}, it does lead to a significant improvement in MARE. The MMCBE dataset was captured in very flat and relatively texture-less fields and this result demonstrates that explicit geometric supervision ($\mathcal{L}_{g}$) in NeFF is critical to biomass prediction in this context. We next tested whether the use of structural information (features) or geometrical coordinates (xyz) as input into BioNet yielded greater prediction accuracy. As seen in~\cref{tab:feat_depth_sup}, 3D features indeed improved MARE by 2.1\%, suggesting that semantic features help BioNet to distinguish plants from foreground and background. Feature visualization can be found in~\cref{fig:3d_feature_map}. Finally, we conducted an ablation study on the number of Transformer encoders in BioNet (~\cref{fig:transformer_encoder}). The `zero' scenario corresponds to the absence of Transformers in the BioNet model. In this case, the output of the sparse 3D backbone network is directly connected with the last two layers of the MLP for biomass prediction. The results of this experiment show that the prediction error decreases with $Num$ from 1, then increases slightly after $Num \ge 5$, and we used 5 Transformer encoders for the final biomass prediction. 

\section{Conclusion}
\label{sec:conclusion}

\noindent To estimate biomass in an accurate and scalable manner, we introduced a novel biomass prediction network, Generalized BioNet, that can process both image and point cloud data. We evaluated this network using two public datasets. Experimental results show that our method outperforms SOTA regarding prediction accuracy. The vision-based prediction solution can facilitate the measurement of an important trait in plant breeding at scale. In the future, the collection of additional datasets with both RGB images and LiDAR points will be used to test the generalization of this model across different years and crop varieties. 


{\small
\bibliographystyle{unsrt}
\bibliography{egbib}
}
\end{document}


\maketitle

\begin{figure}[h]
    \centering
    \includegraphics[width=1.0\linewidth]{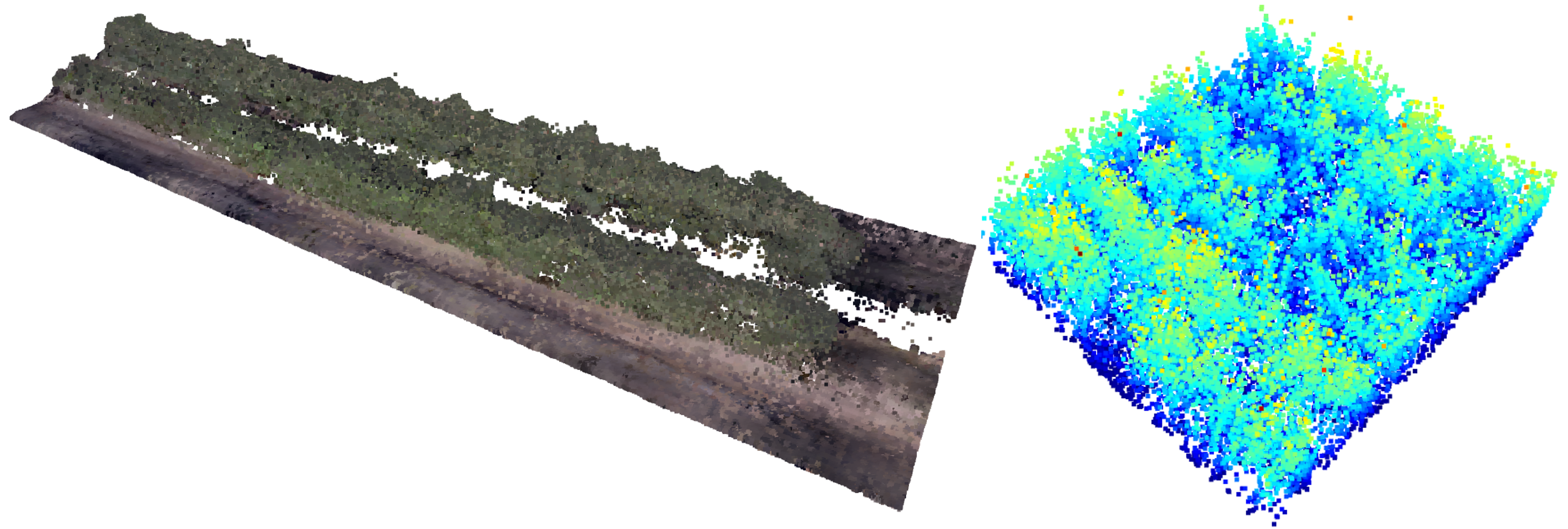}
    \caption{Visualization of a point cloud from the MMCBE (left) and wheat (right) datasets. This figure comes from~\cite{li2024mmcbe}.}
    \label{fig:pc_vis}
\end{figure}

\section{Implementation}
\begin{figure}[h]
    \centering
    \includegraphics[width=1.0\columnwidth]{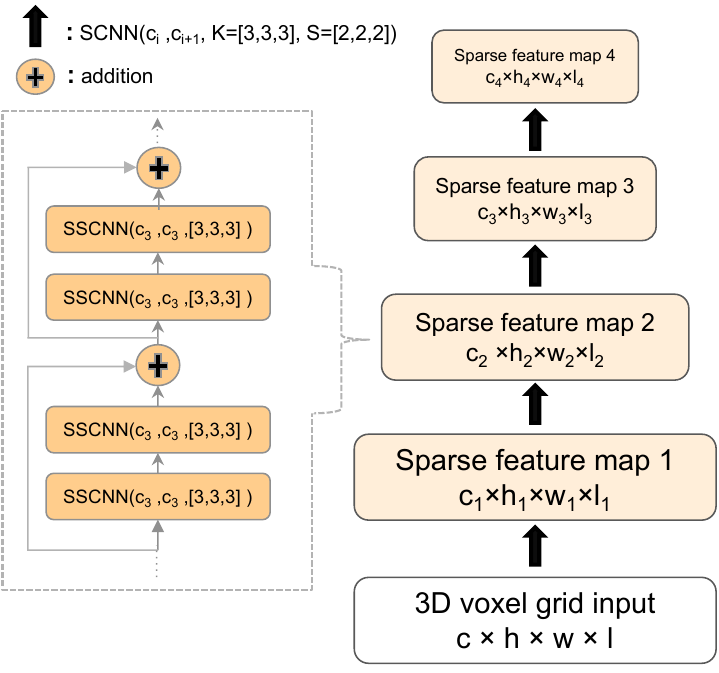}
    \caption{Architecture of the sparse 3D CNN backbone. The input is the voxelized 3D features generated from NeFF. SSCNN stands for submanifold sparse 3D CNN\cite{graham2017submanifold, li2019three}, while SCNN stands for sparse 3D CNN. Each pyramidal level consists of two ResNet blocks \cite{he2016identity}.}
    \label{fig:3dbn}
\end{figure}

In our NeFF, the SDF network $\mathcal{E}_{g}$ is modeled by an MLP built by 8 hidden layers, and Relu with Softplus is used as the activation function for all hidden layers. A skip connection is used to connect the input and the fourth layer. The feature filed $\mathcal{E}_{f}$ consists of an MLP with two hidden layers, while the color filed $\mathcal{E}_{c}$ has four hidden layers. All hidden layers have the same hidden size of 256. Positional encoding is applied to spatial locations with 6 frequencies and viewing directions with 4 frequencies. We use 4 sparse 3D CNN blocks, as shown in ~\cref{fig:3dbn}, to compress the 3D features map into a 2D one, and the 3 Transformer encoders have the same defaulting as the original paper \cite{vaswani2017attention}. The final prediction MLP includes two hidden layers with 512 and 256 hidden sizes. We use an ADAM optimizer \cite{kingma2014adam} to optimize both networks on a Navida A5500 GPU. The $\mathcal{F}$ and $\mathcal{H}$ functions are trained separately. It takes 10 hours to optimize NeFF for 200K iterations, and around 8 hours to train BioNet for 100K iterations with 4 batch sizes. 


\section{Difference in two datasets}
When comparing our method with baselines in Section 5.2, we find that the MARE suggests that the same approach achieves superior results on the wheat dataset~\cite{pan2022biomass} compared to MMCBE~\cite{li2024mmcbe}. To delve deeper into these differences, we visualized the point clouds from both datasets, as shown in Figure~\ref{fig:pc_vis}. Significant occlusion issues exist in MMCBE, where LiDAR data is often scanned from the side, contrasting with the top-down collection approach in the wheat dataset. This difference can lead to a higher quality of point cloud data for the wheat dataset, thereby enhancing the accuracy of biomass prediction.

\section{3D backbone network architecture}
\label{sec:3dbk}
We are using the 3D backbone network to further extract 3D features from the point cloud or distilled feature maps, and the 3D backbone network mainly consists of sparse 3D CNN layers due to their advantages in efficient computation and local geometrical feature extraction~\cite{li2020object, s18103337}. The details of 3D backbone network are shown in~\cref{fig:3dbn}. For the MMCBE dataset, we first voxelized the inputs (i.e. point cloud or 3D feature maps) to obtain its voxel representation. The size of the 3D inputs is $16$ meters in length, $4$ meters in width, and $1.5$ meters in height, with a voxelization resolution of $0.08$ meters in length, $0.05$ meters in width, and $0.075$ meters in height, respectively. For the wheat dataset, the size of the 3D input is 1 $m^3$, the voxelization resolution is 0.008 meters in width and length, and 0.025 meters in height. The value of each voxel is obtained by averaging all points inside a given voxel. The computational operations mainly include regular $3 \times 3\times 3$ convolutional kernel, submanifold CNN \cite{graham2017submanifold} and max pooling.


\begin{figure*}[t!]
\begin{subfigure}{.5\textwidth}
  \centering
  \includegraphics[width=1.0\linewidth]{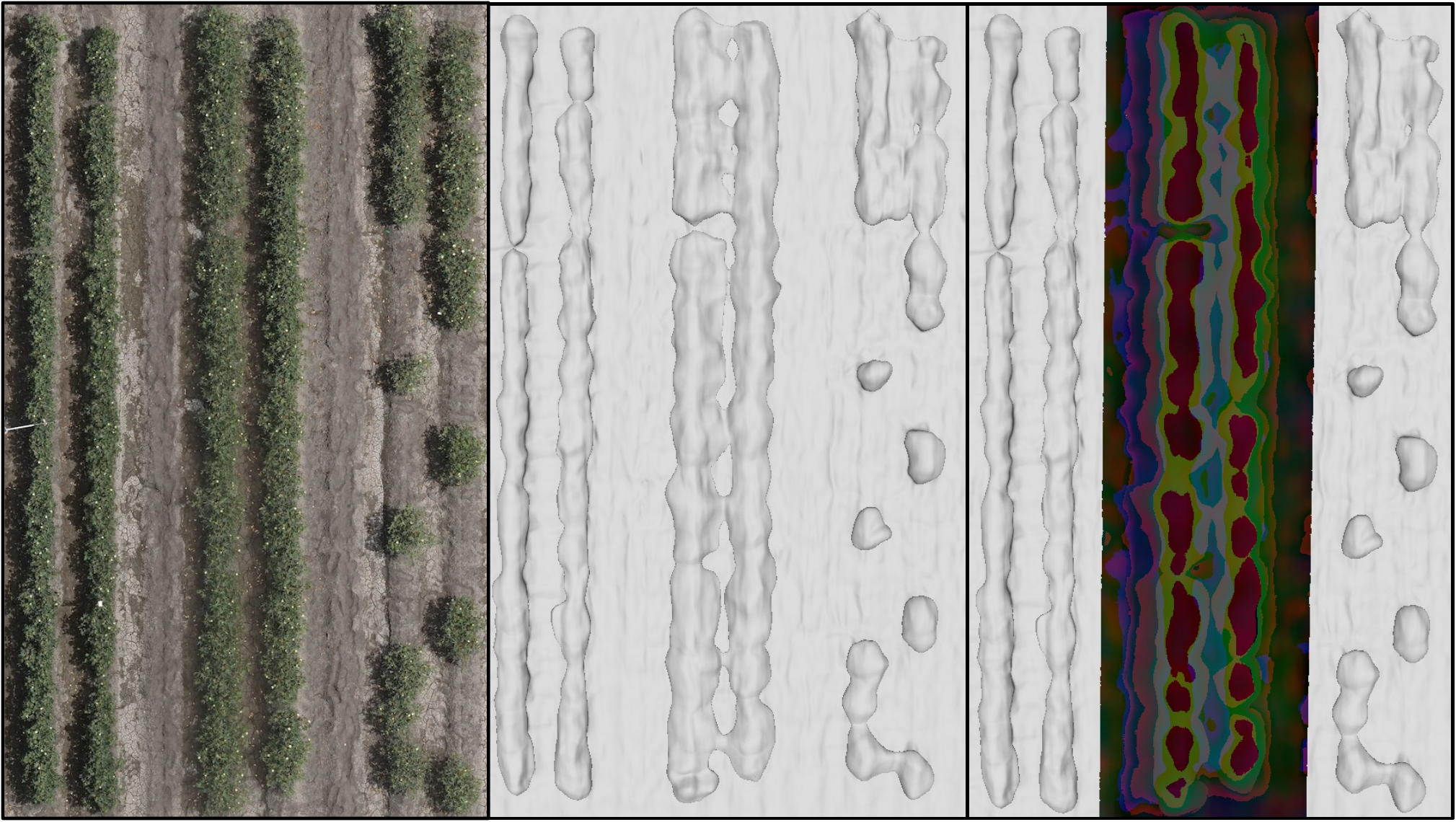}
  \caption{ }
\end{subfigure}
\begin{subfigure}{.5\textwidth}
  \centering
  \includegraphics[width=1.0\linewidth]{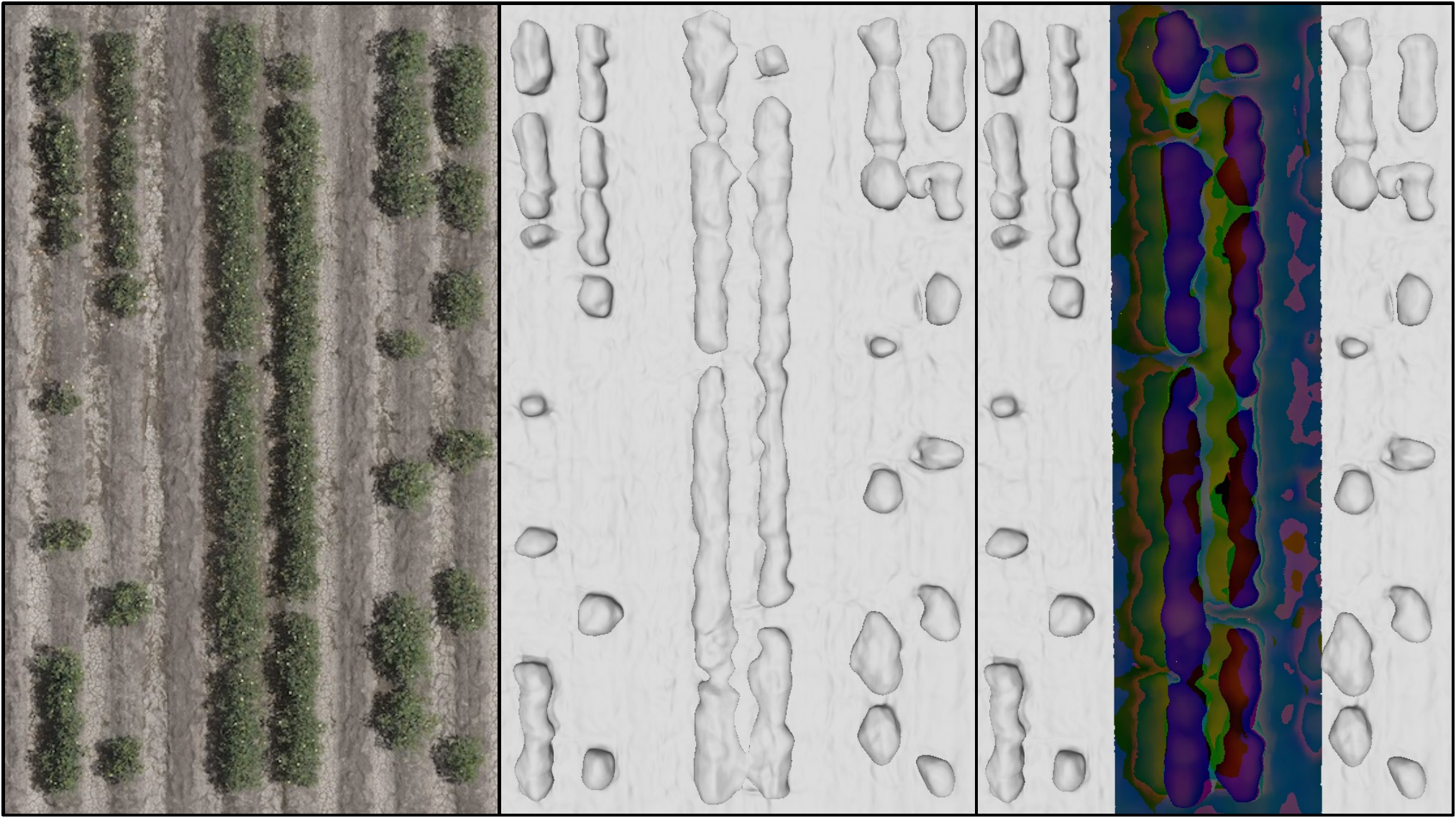}
  \caption{ }
\end{subfigure}
\hfill
\begin{subfigure}{.5\textwidth}
    \centering
    \includegraphics[width=1.0\linewidth]{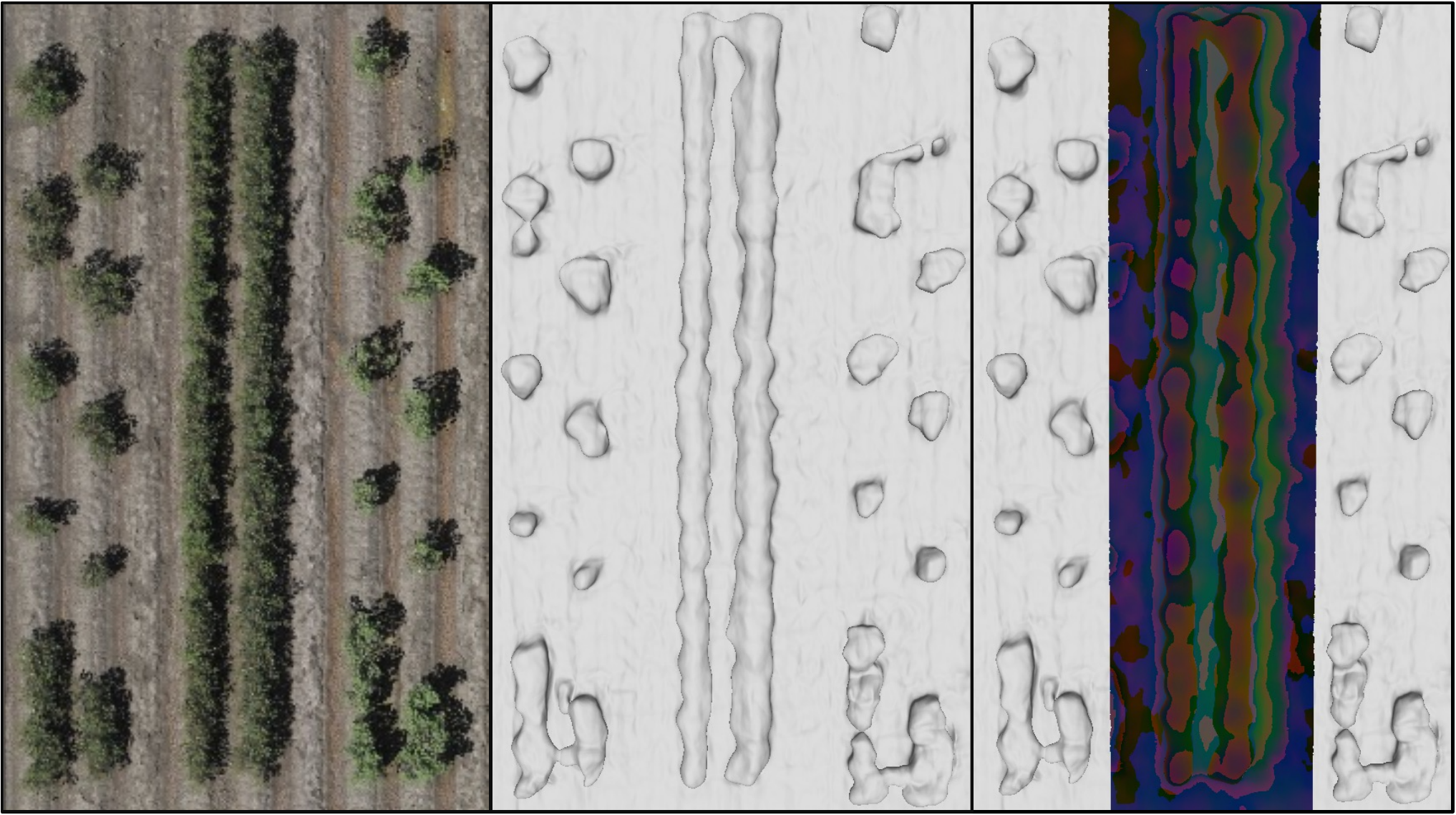}
    \caption{ }
\end{subfigure}
\begin{subfigure}{.495\textwidth}
    \centering
    \includegraphics[width=1.0\linewidth]{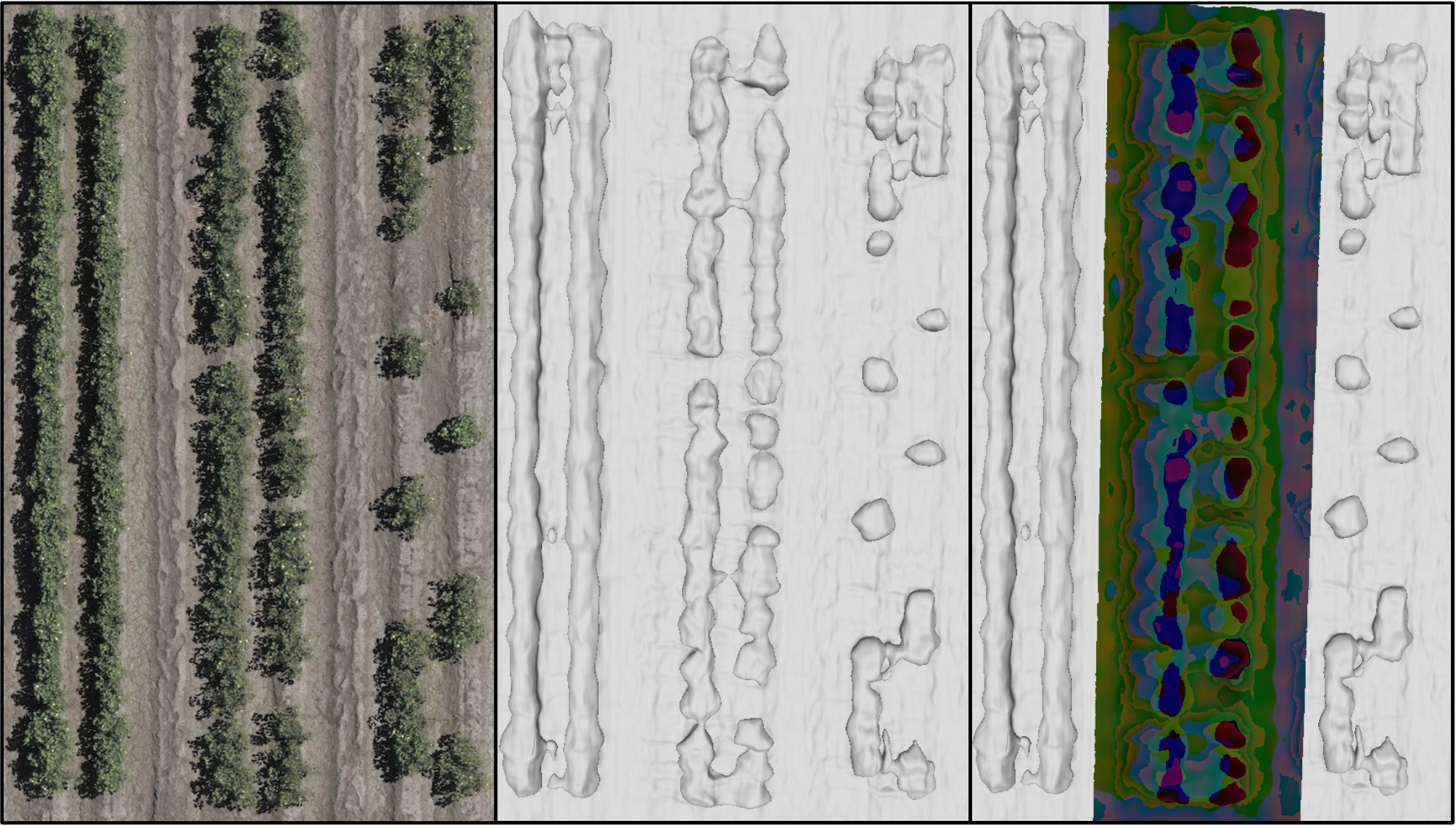}
    \caption{ }
\end{subfigure}
\caption{The visualization of 3D features generated by NeFF module. For each fused image, the stitched RGB image is on the leftmost, a 3D reconstruction is in the middle, and the rightmost image shows cropped 3D features (used by the BioNet module for biomass prediction.)}
\label{fig:3d_features}
\end{figure*}

\section{Evaluation metrics} 
Metrics used in the paper are namely mean absolute error \textit{(MAE)}, mean absolute relative error (\textit{MARE}), and root mean square error \textit{(RMSE}), as shown in equation~\ref{equ:mae}~\ref{equ:mare}~\ref{equ:rmse}. \textit{MARE} offers the advantage of being more robust against the range of ground truth values observed across our 9 timepoints.
\begin{equation}
    \text{MAE} = \frac{1}{N} \sum_{i=1}^{N} |m_i - \hat{m}_i|
    \label{equ:mae}
\end{equation}

\begin{equation}
    \text{MARE} = \frac{1}{N} \sum_{i=1}^{N} \frac{|m_i - \hat{m}_i|}{m_i}
    \label{equ:mare}
\end{equation}

\begin{equation}
    \text{RMSE} = \sqrt{\frac{1}{N} \sum_{i=1}^{N} (m_i - \hat{m}_i)^2}
    \label{equ:rmse}
\end{equation}

Where  \( N \) is the total point cloud data points with \( i \in \{1, \ldots, N\} \) as the index, \( \hat{m}_i \) is the predicted and \(m_i \) the ground-truth biomass value. The lower these errors, the better the prediction accuracy. We used RI to indicate the relative improvement when comparing two different methods. The accuracy of two methods are $p1$ and $p2$, (say $p1$ is better than $p2$), the R1 of $p1$ against $p2$ is as follows:
\begin{equation}
RI = \frac{|p1 - p2|}{p2}
\end{equation}

\section{NeFF}
\label{sec:neff}
To visualize the results from the NeFF, we sample 3D features at a resolution of $1024^3$ from the neural feature field (the feature sampling resolution used for BioNet module in $2048^3$) in MMCBE dataset. We then employ principal component analysis to compress the high-dimensional features into three channels. These three channels are subsequently normalized to the range [0, 1] and displayed as the RGB color of reconstructed points, as shown in~\cref{fig:3d_features}. From this visualization, it is evident that the distilled 3D features can effectively distinguish between the foreground (plants) and background (soil). Furthermore, different parts of the plants can be separated as well. The NeFF generally constructs 3D features for each plot, encompassing several rows of cotton. However, our specific interest lies in the two 13 ${m}$ cotton rows, as we possess above-ground biomass ground truth data only for every two rows of cotton. Consequently, we must crop the 3D features of the two cotton rows of interest from the entire set of constructed features. These cropped 3D features serve as the input for the BioNet module, which predicts the final biomass. The cropped 3D features are depicted in each rightmost figure in Fig. \ref{fig:3d_features}.

{\small
\bibliographystyle{unsrt}
\bibliography{egbib}
}